%% file: ijcai26.tex
\documentclass{article}
\usepackage{ijcai26}

\usepackage{times}
\usepackage{soul}
\usepackage{url}
\usepackage[hidelinks]{hyperref}
\usepackage[utf8]{inputenc}
\usepackage[small]{caption}
\usepackage{graphicx}
\usepackage{amsmath}
\usepackage{amsthm}
\usepackage{booktabs}
\usepackage{algorithmic}
\usepackage{algorithm}
\usepackage[switch]{lineno}
\usepackage{amssymb}
\usepackage{multirow}
\usepackage{xcolor}
\usepackage[table,xcdraw]{xcolor}
\usepackage{natbib}
\newcommand{\methodname}{\texttt{OJBKQ}}

\title{\methodname: \textbf{O}bjective-\textbf{J}oint \textbf{B}abai-Klein \textbf{Q}uantization}

\author{
Xinyu Wang$^{1,*}$
\and
Ziyu Zhao $^{1,*}$\and
Peng Lu$^2$\and
Yu Gu$^1$ \and
Xiao-Wen Chang$^1$\\
\affiliations
$^1$McGill University\\
$^2$Université de Montréal \\
$^*$Equal contribution\\
\emails
\{xinyu.wang5, ziyu.zhao2\}@mail.mcgill.ca,
chang@cs.mcgill.ca,
}

\begin{document}

\maketitle

\input{sections/0_abstract}
\input{sections/1_intro}
\input{sections/2_related_work}
\input{sections/3_methodology}
\input{sections/4_experiment}
\input{sections/5_conclusion}

\bibliographystyle{named}
\bibliography{ijcai26}
\clearpage
\appendix
\input{sections/x_appendix}

\end{document}

%% file: sections/0_abstract.tex
\begin{abstract}
    Post-training quantization (PTQ) is widely used to compress large language models without retraining. However, many existing weight-only methods rely on heuristic objectives and greedy rounding, thus leading to noticeable degradation under low-bit quantization In this work, we introduce OJBKQ (Objective-Joint Babai-Klein Quantization with K-Best Sampling), a layer-wise PTQ method that formulates weight quantization as a joint optimization problem over activations and weights. This formulation results in 
    a multiple-right-hand-side box-constrained integer least squares (BILS) problem in each layer, which is NP-hard. 
    For each column of the weight matrix, we apply an extended Babai nearest-plane algorithm and an extended version of Klein’s randomized Babai algorithm to find the minimum-residual Babai–Klein point, a sub-optimal solution to the BILS problem.  
    Experimental results on large language models show that OJBKQ achieves lower perplexity at 3–4 bits compared to existing PTQ approaches, while maintaining comparable computational cost. 
\end{abstract}

%% file: sections/1_intro.tex
\section{Introduction}

Large Language Models (LLMs) \citep{achiam2023gpt,touvron2023llama,comanici2025gemini25pushingfrontier} 
pose substantial deployment challenges due to their massive memory footprint and inference costs \citep{chen2023frugalgpt}.
Layer-wise post-training quantization (PTQ) \citep{frantar2023gptqaccurateposttrainingquantization,awq,chee2024quip2bitquantizationlarge} has become a widely adopted compression standard, enabling efficient deployment by sequentially quantizing parameters without retraining.
In this work, we focus on standard and practically relevant 4-bit and 3-bit settings (e.g., group size 128) and study how to further improve quantization quality beyond strong PTQ baselines.

Despite its empirical success, layer-wise PTQ is intrinsically a \emph{discrete} optimization problem.
At its core, each layer-wise step solves an integer-constrained quadratic subproblem, typically approached with sequential rounding-type solvers.
Recent theoretical analyses \citep{chen2025geometryllmquantizationgptq} reveal that standard solvers (e.g., GPTQ) admit a lattice-decoding interpretation: they are closely related to Babai’s nearest-plane algorithm \citep{babai1986lovasz}, which finds the Babai point,
a suboptimal solution to the Closest Vector Problem (CVP)\citep{agrell2002closest}, which is also referred to as the Integer Least-Squares (ILS) problem. 
However, when the lattice basis matrix is not well-conditioned
or when the dimension is large, the Babai point residual
may be much larger than the optimal residual.
Thus, we need to find a better sub-optimal solution.
% This perspective highlights two orthogonal failure modes in the PTQ pipeline: \emph{how we generate candidates} and \emph{how we select among them}.
% % First, existing solvers typically follow a \emph{single, deterministic} decoding trajectory. In the context of CVP, this greedy approach can be substantially suboptimal when the induced lattice basis (from triangular factors of the Hessian) is highly non-orthogonal (large off-diagonal couplings in the triangular factor). \peng{This should be a problem for other applications as well if you wanna apply CVP solvers. You can leverage Prof. Chang's expertise here to add some references which mentioned the some issues when applying CVP to other domains.}
% This suggests that lightweight randomized multi-trial decoding could uncover superior integer solutions at a modest computational cost. \peng{Cite some related works to support this suggestion.}
Second, minimizing a layer-local proxy loss does not necessarily translate to better end-to-end quantization due to error propagation and input distribution drift; consequently, the \emph{selection objective} used to compare and choose among candidates is as crucial as the solver itself, yet remains inconsistent across methods.

We propose \textbf{\methodname} (Objective-Joint Babai-Klein Quantization), a unified framework  that addresses both issues.
 \textbf{(i) Joint Target Alignment (JTA):} To better align layer-wise decisions with end-to-end behavior under error propagation, we introduce a unified candidate \emph{scoring} objective (Eq.~\ref{eq:jta_score}) with a single continuous knob.
JTA interpolates between two fundamental alignment targets: matching the \emph{runtime} quantized activations induced by the partially-quantized network, and matching the corresponding \emph{full-precision reference} outputs \citep{arai2025quantization}.
The formulated quantization problems at each layer are
multiple-right-hand-side box-constrained integer least squares (BILS) problems.
\textbf{(ii) Random-$K$ quantization:} 
For each column of the weight matrix in a BILS problem,
we apply the box-constrained Babai algorithm to obtain one sub-optimal solution 
(a low-bit vector candidate).
For the same column, we then extend Klein's K-time randomized Babai algorithm \citep{10.5555/338219.338661}
to the box-constrained case to generate $K$ additional sub-optimal solutions 
independently.
Among these $K+1$ candidates, we select the one which has the minimum
residual, referred to as the best Babai-Klein point. 
There are two levels of parallelization. One across the columns of the weight matrix and the other across the K points produced by the box-constrained 
Klein algorithm. Both can be implemented efficiently. 

In \methodname, Random-$K$ expands the quality of the \emph{candidate set}, while JTA improves \emph{how we select} the optimal candidate.

Our contributions are summarized as follows:
\begin{itemize}
    \item \textbf{Quantization Formulation.} We revisit layer-wise PTQ from a CVP perspective and reformulate it as an explicit lattice-decoding problem. This enables a faithful Babai-style implementation that operates via triangular factors and back-substitution, without materializing any matrix inverse.
    \item \textbf{Superior Sub-optimal Solution.} 
    We extend Klein's randomized algorithm
    to the box-constrained case and apply it to PTQ. By running $K$ independent probabilistic traces in parallel, our method explores non-orthogonal lattice bases to find superior integer candidates while maintaining strong language modeling performance with low perplexity in both 4‑bit and 3‑bit configurations.
    In designing and implementation of our numerical algorithm, we take both accuracy 
    and efficiency into account. For example, unlike GPTQ, no inverse of a matrix
    is computed. To reduce the computational overhead of multiple indepandent solution, we design a GPU-efficient algorithm presented in Appendix \ref{app}

    \item \textbf{JTA Candidate Selection.} We propose the \textbf{Joint Target Alignment (JTA)} objective, a unified scoring criterion that subsumes commonly used PTQ objectives as special cases by varying the alignment target between runtime-quantized and full-precision references, providing a more fine-grained criterion for candidate selection under error propagation.
\end{itemize}

%% file: sections/2_related_work.tex
\section{Related Work}

\paragraph{Layer-wise post-training quantization (PTQ).}
Post-training quantization is a practical standard for compressing large language models without retraining.
Following recent PTQ taxonomies \citep{zhao2025benchmarkingposttrainingquantizationllms}, mainstream weight-only PTQ methods can be broadly grouped by their dominant mechanism.
\emph{Compensation-based} approaches, exemplified by GPTQ \citep{frantar2023gptqaccurateposttrainingquantization}, quantize weights sequentially while updating the remaining unquantized weights to compensate for accumulated quantization errors, often leveraging second-order curvature surrogates estimated from calibration data.
\emph{Rotation-based} methods such as QuIP \citep{chee2024quip2bitquantizationlarge} apply structured transformations to reshape weight distributions and improve quantizability under low-bit constraints.
\emph{Salience-based} methods, represented by AWQ \citep{awq}, reduce quantization error by selecting scaling factors according to activation/weight importance, avoiding mixed-precision deployment.
In addition, \emph{optimization-based} approaches (e.g., OmniQuant\footnote{Although it is classified as PTQ in the survey, it still requires training on the calibration dataset to find scales.Then we exclude it from our baselines}and PoTPTQ)  \citep{shao2024omniquant,wang2025potptqtwosteppoweroftwoposttraining} directly optimize quantization parameters with lightweight objectives while keeping pretrained weights frozen.
Overall, existing PTQ pipelines mainly differ along two orthogonal axes: (i) the approximate \emph{solver} used to handle the discrete subproblem, and (ii) the \emph{objective} used to evaluate candidates and guide selection.

\paragraph{Integer least squares and Babai-type solution.}
The integer least squares (ILS) problems or/and its box-constrained variant (BILS) arise in crypotograph \citep{MicG02}, GPS \citep{Teu96,ChaYZ05,10206541}, MIMO detection \citep{AgrEVZ02,BaiCY14} etc, where the goal is to recover a discrete-valued vector under a quadratic objective.
Since the ILS and BILS are NP-hard \citep{micciancio2002hardness,Ver89},
in some application often efficient sub-optimal solutions are applied.
An often used method is Babai’s nearest-plane algorithm \citep{babai1986lovasz}, which computes the Babai point by back substitution on an upper-triangular 
linear system, where rounding is performed at each step. 
The Babai point is the first integer point found by the 
Schnorr–Euchner sphere-decoding algorithm, which enumerates integer points
in an ellipsoid to find the optimal solution \citep{SchE94}.
It has been extended to the box-constrained case \citep{WenChang2021BILS}.
While computationally efficient, when the coefficient matrix (referred to 
as the lattice basis matrix in lattice theory) is not well-conditioned
(or not well-reduced in lattice theory),
or when the dimension of the integer vector is large. 
% the dimension is not large, typically smaller than 100), 
% the Babai point may have much larger residual than the optimal solution.
% For the former, one could apply the famous LLL reduction \citep{LenLL82}
% to mitigate it, but in quantization, the dimension is large, applying 
% LLL is computationally expensive.
There are other sub-optimal algorithms, such as those proposed in \citep{barbero2006performance,chang2024extended},
but they are not suitable for GPU computing. 
Fortunately, there is a randomized version of the Babai algorithm 
proposed in \citep{10.5555/338219.338661}, which is not widely used yet.
In each step of the Babai algorithm, Klein's algorithm rounds
the real solution to a nearby integer according to a predefinded distribution. 
When the algorithm is called $K$ times, it generates $K$ points. 
Then the best one, which gives the minimum residual, is chosen as the final sub-optimal
solution. This method  can be parallelized.
The method is for unconstrained ILS problem and has not been widely used yet. 
Recently, the connection between sequential PTQ solvers and the Babai point has gained attention\citep{chen2025geometryllmquantizationgptq}.
This paper emphasizes that the PTQ problems can be modeled as BILS problems so that sub-optimal BILS algorithms can be applied. 

\paragraph{Objectives under error propagation and distribution drift.}
A recurring challenge in layer-wise PTQ is that minimizing a layer-local reconstruction proxy does not necessarily translate to improved end-to-end quality, due to quantization error propagation and the resulting input distribution drift across layers.
Consequently, the choice of the \emph{alignment target}---whether to match outputs under runtime (partially-quantized) activations or to match full-precision reference outputs---plays a central role in practical PTQ performance.
Recent work revisits this issue from an error-propagation viewpoint and advocates objectives that better reflect downstream behavior under partial quantization \citep{arai2025quantization}.
Our JTA objective follows this perspective but provides a unified scoring criterion with a single continuous knob that interpolates between runtime-quantized activations and full-precision reference targets, enabling consistent candidate selection across layers.

%% file: sections/3_methodology.tex
\newcommand{\best}[1]{\textbf{#1}}
\newcommand{\second}[1]{\underline{#1}}

\section{Methodology}
\label{sec:method}

\subsection{Layer-wise Objectives and Joint Target Alignment (JTA) }
\label{subsec:jta}

We now start analyze commonly used layer-wise PTQ objectives.
Consider a linear layer with full-precision weight matrix $\mathbf{W}\in\mathbb{R}^{m \times n}$. 
Let $\mathbf{X}\in\mathbb{R}^{p \times m}$ be full-precision calibration activations,  
$\widetilde{\mathbf{X}}\in\mathbb{R}^{p \times m}$ be the corresponding \emph{runtime} activations produced by a partially quantized network (i.e., upstream layers already quantized), $\mathbf{Y}^{\mathrm{fp}} := \mathbf{XW}$ denote the full-precision reference output, and
$\mathbf{Y}^{\mathrm{rt}} := \widetilde{\mathbf{X}}\mathbf{W}$ denote the runtime-consistent reference output under partially quantized activations.
At runtime, a candidate $\widehat{\mathbf{W}}$ produces the output $\widehat{\mathbf{Y}} := \widetilde{\mathbf{X}}\widehat{\mathbf{W}}$.

Many existing objectives can be interpreted as choosing different alignment targets:
\begin{align}
\small
\label{eq:obj_rt}
\text{(Runtime-consistent)}\quad 
&\min_{\widehat{\mathbf{W}}} \;\;
\|\widetilde{\mathbf{X}}\widehat{\mathbf{W}} - \widetilde{\mathbf{X}}\mathbf{W}\|_F^2 \\
&\qquad = \|\widehat{\mathbf{Y}}-\mathbf{Y}^{\mathrm{rt}}\|_F^2, \\
\label{eq:obj_fp}
\text{(Full-precision mapping)}\quad 
&\min_{\widehat{\mathbf{W}}} \;\;
\|\mathbf{X}\widehat{\mathbf{W}} - \mathbf{X}\mathbf{W}\|_F^2, \\
\label{eq:obj_mismatch}
\text{(Mismatch target)}\quad 
&\min_{\widehat{\mathbf{W}}} \;\;
\|\widetilde{\mathbf{X}}\widehat{\mathbf{W}} - \mathbf{X}\mathbf{W}\|_F^2 \\
&\qquad = \|\widehat{\mathbf{Y}}-\mathbf{Y}^{\mathrm{fp}}\|_F^2.
\end{align}
Eq.~\eqref{eq:obj_rt} directly reflects runtime behavior but aligns to a target computed on quantized activations, which may be noisy; this objective is adopted by GPTQ and QUIP.
Eq.~\eqref{eq:obj_fp} admits a convenient least-squares structure (prior to discretization) but suffers from train--run mismatch when earlier layers are quantized, and is optimized by AWQ.
Eq.~\eqref{eq:obj_mismatch} partially alleviates this mismatch by evaluating candidates on runtime activations while preserving the full-precision reference; QEP employs this objective as a corrective patch within a layer-wise quantization pipeline, although it may over-penalize candidates when error drift accumulates across layers.

These trade-offs motivate a unified selection objective that interpolates between runtime-consistent and full-precision references, while optionally controlling weight drift.

We propose an expressive \emph{Joint Target Alignment (JTA)} objective with a single continuous knob to score and select candidates generated by Random-$K$ decoding.
We define an interpolated target
\begin{equation}
\small
\label{eq:jta_target}
\mathbf{Y}^\star(\mu) := (1-\mu)\,\mathbf{X}\mathbf{W} + \mu\,\widetilde{\mathbf{X}}\mathbf{W},
\qquad \mu\in[0,1],
\end{equation}
and formulate the ILS problem: 
\begin{equation}
\label{eq:jta_score}
\min\mathcal{S}(\widehat{\mathbf{W}})
\;:=\;
\|\widetilde{\mathbf{X}}\widehat{\mathbf{W}}-\mathbf{Y}^\star(\mu)\|_F^2
\;+\;
\lambda^2\|\widehat{\mathbf{W}}-\mathbf{W}\|_F^2,
\end{equation}
where $\widehat{\mathbf{W}}$ is subject to some constraint.
Later we will show this is a multiple-right-hand-side box-constrained integer least squares (BILS) problem and provide a sub-optimal solution.  
The knob $\mu$ interpolates between matching full-precision and runtime-consistent references, while $\lambda$ optionally regularizes weight displacement.
In our framework, \textbf{Random-$K$ quantization} (Sec.~\ref{subsec:randomk}) 
rewrites find a sub-optimal solution to , and \textbf{JTA} (Eq.~\eqref{eq:jta_score}) provides a consistent curvature-aware criterion for selecting the best candidate under error propagation.
When $\mu{=}1$ and $\lambda{=}0$, Eq.~\eqref{eq:jta_score} reduces to the runtime-consistent objective in Eq.~\eqref{eq:obj_rt}; when $\mu{=}0$ and $\lambda{=}0$, it aligns to the full-precision reference in Eq.~\eqref{eq:obj_mismatch}.

\paragraph{End-to-end layer-wise procedure.}
For each layer, we (i) form $\widetilde{\mathbf{X}}$ by a partially quantized forward pass, (ii) find a sub-optimal optimal solution to Eq.~\eqref{eq:jta_score}, 
to be seen in the next sections.
This closes the loop between discrete candidate generation and objective-aligned selection.
% ------------------------------------------------------------

% \subsection{Integer Least Squares and Babai-type Decoding}
% \label{subsec:ils}
% \textcolor{red}{{XW: Will see how to handle this subsection and the next one later.}}
% We begin with the classical Integer Least Squares (ILS) problem:
% \begin{equation}
% \label{eq:ILS}
%     \min_{\mathbf{z}\in \mathbb{Z}^n}\;\; \|\mathbf{y}-\mathbf{A}\mathbf{z}\|_2^2,
% \end{equation}
% where $\mathbf{y}\in\mathbb{R}^m$ is a target vector and $\mathbf{A}\in\mathbb{R}^{m\times n}$ has full column rank.
% Solving Eq.~\eqref{eq:ILS} exactly is NP-hard in general~\citep{micciancio2002hardness}.
% When $\mathbf{z}$ is additionally constrained to a bounded integer set, Eq.~\eqref{eq:ILS} becomes a \emph{box-constrained} ILS (BILS).
% A widely used efficient approximation is the \emph{Babai point}, obtained by Babai's nearest-plane algorithm~\citep{babai1986lovasz}, which performs greedy rounding along a triangular factorization.
% Randomized variants (e.g., Klein-type sampling\peng{cite their paper.}) draw multiple independent decoding traces and return the best candidate (best-of-$K$), often improving solution quality when the induced basis is highly non-orthogonal.

% Recent analyses connect layer-wise PTQ solvers to lattice decoding and ILS/BILS formulations~\citep{chen2025geometryllmquantizationgptq}.
% We next formalize layer-wise quantization as a structured BILS and design a GPU-efficient randomized $K$-best Babai solver, paired with a unified candidate scoring objective.

\subsection{From Layer-wise Quantization to Integer Least Squares}
\label{subsec:ils_bridge}

From \eqref{eq:jta_score} we obtain 
\begin{equation}
\mathcal{S}(\widehat{\mathbf{W}})
\;:=\;
\left\|\begin{bmatrix} \widetilde{\mathbf{X}} \\ \lambda \mathbf{I}  
\end{bmatrix}\widehat{\mathbf{W}}
-\begin{bmatrix} \mathbf{Y}^\star(\mu) \\ \lambda\mathbf{W}\end{bmatrix}\right\|_F^2.
\label{eq:new_obj}
\end{equation}
The task of finding an optimal quantized weight $\widehat{\mathbf{W}}$ to minimize the selection objective in Sec.~\ref{subsec:jta} can be formally mapped to the classical \emph{Integer Least Squares} (ILS) problem. 
\begin{figure}
    \centering
    \includegraphics[width=\linewidth]{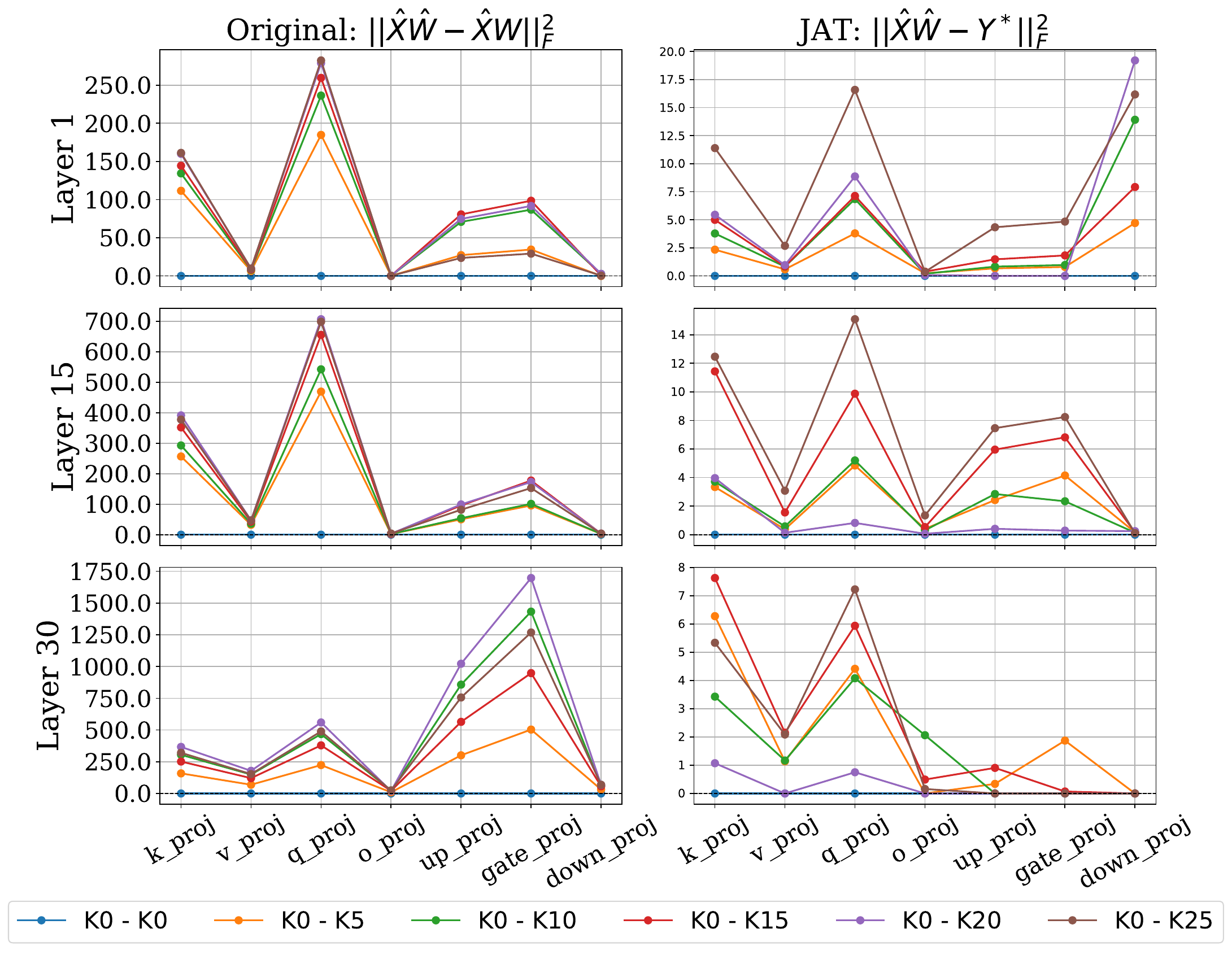}
    \caption{Layer-wise comparison of the original output norms and JTA reconstruction errors across Layers 1, 15, and 30.We present results for all linear modules under varying $K$ settings.}
    \label{fig:Residual_over_K}
\end{figure}
To facilitate the derivation, we first introduce additional notation. Let
\[
\mathbb{B} = \{0, 1, \ldots, 2^{\text{wbit}} - 1\}
\]
denote the box constraint corresponding to the set of admissible integer values for a $\text{wbit}$-bit quantized weight. Let
\[
\mathbf{Q} = [\mathbf{q}_1, \mathbf{q}_2, \ldots, \mathbf{q}_n] \in \mathbb{B}^{m \times n}
\]
denote the quantized version of the full-precision weight matrix $\mathbf{W}$. We further define
\[
\mathbf{S} = [\mathbf{s}_1, \mathbf{s}_2, \ldots, \mathbf{s}_n] \in \mathbb{R}^{m \times n}, \quad
\mathbf{Z} = [\mathbf{z}_1, \mathbf{z}_2, \ldots, \mathbf{z}_n] \in \mathbb{R}^{m \times n}
\]
as the scale matrix and zero-point matrix, respectively. Both $\mathbf{S}$ and $\mathbf{Z}$ are pre-computed using standard statistical calibration methods (e.g., the Absmax method).

Finally, the dequantized weight matrix is given by
\begin{equation*}
\widehat{\mathbf{W}} = \mathbf{S} \odot (\mathbf{Q} - \mathbf{Z}),
\end{equation*}
where $\odot$ denotes element-wise multiplication. The matrices $\mathbf{Q}$, $\mathbf{S}$, $\mathbf{Z}$, and $\widehat{\mathbf{W}}$ are all partitioned column-wise in the same manner as $\mathbf{W}$.
% Then we define $\mathbf{A}:=\begin{bmatrix} \widetilde{\mathbf{X}} \\ \lambda \mathbf{I}  
% \end{bmatrix}$  and $\mathbf{T}=\begin{bmatrix} \mathbf{Y}^\star(\mu) \\ \lambda\mathbf{W}\end{bmatrix}$.  With above notation, we can rewrite \eqref{eq:new_obj}  to 
% \begin{equation}
% \sum_{j=1}^n \|\mathbf{A}\mathbf{D}_j\mathbf{q}_j - \mathbf{b}_j \|_2^2\end{equation}
% where $\mathbf{D}_j=\text{diag} (\mathbf{s}_j) \in \mathbb{R}^{m\times m}$
% and $\mathbf{b}_j=\mathbf{t}_j+\mathbf{A}\mathbf{D}_j\mathbf{z}_j$. As a result, Eq \ref{eq:new_obj} becomes $n$ independant box-constrained integer least squares problems
% \begin{equation}
%  \label{eq:n_opt}
% \min_{\mathbf{q}_j\in \mathbb{B}^m}\| \mathbf{A}\mathbf{D}_j\mathbf{q}_j - \mathbf{b}_j\|_2^2, \ \ 
% j=1,\ldots,n.
% \end{equation}
% As each problem is independent and to simplify the notation, we only consider a per-column sub-problem as 
% \begin{equation}
%  \label{eq:opt}
% \min_{\mathbf{q}\in \mathbb{B}^m}\| \mathbf{A}\mathbf{D}\mathbf{q} - \mathbf{b}\|_2^2, \ \ 
% \end{equation}
% Solving Eq.~\eqref{eq:opt} exactly is NP-hard for general lattices~\citep{micciancio2002hardness}.Specifically, the \emph{Babai point}~\citep{babai1986lovasz} provides an efficient approximation by performing successive orthogonal projections and greedy rounding along a triangular factorization of $\mathbf{A}$.However, recent analyses~\citep{chen2025geometryllmquantizationgptq} suggest that layer-wise solvers like GPTQ can be interpreted as greedy approximations to this lattice decoding problem with different objective. but the idea is the same
We further define
\[
\mathbf{A} := 
\begin{bmatrix}
\widetilde{\mathbf{X}} \\
\lambda \mathbf{I}
\end{bmatrix},
\qquad
\mathbf{T} :=
\begin{bmatrix}
\mathbf{Y}^\star(\mu) \\
\lambda \mathbf{W}
\end{bmatrix}.
\]
Using the above notation, Eq.~\eqref{eq:new_obj} can be rewritten as
\begin{equation*}
\sum_{j=1}^{n} \left\| \mathbf{A}\mathbf{D}_j \mathbf{q}_j - \mathbf{b}_j \right\|_2^2,
\end{equation*}
where $\mathbf{D}_j = \operatorname{diag}(\mathbf{s}_j) \in \mathbb{R}^{m \times m}$ is a diagonal scaling matrix, and
\[
\mathbf{b}_j = \mathbf{t}_j + \mathbf{A}\mathbf{D}_j \mathbf{z}_j
\]
with $\mathbf{t}_j$ denoting the $j$-th column of $\mathbf{T}$.

As a result, Eq.~\eqref{eq:new_obj} decomposes into $n$ independent box-constrained integer least squares (BILS) problems:
\begin{equation}
\label{eq:n_opt}
\min_{\mathbf{q}_j \in \mathbb{B}^m}
\left\| \mathbf{A}\mathbf{D}_j \mathbf{q}_j - \mathbf{b}_j \right\|_2^2,
\qquad j = 1, \ldots, n.
\end{equation}

Since each subproblem is independent, we simplify notation by focusing on a single per-column subproblem:
\begin{equation}
\label{eq:opt}
\min_{\mathbf{q} \in \mathbb{B}^m}
\left\| \mathbf{A}\mathbf{D}\mathbf{q} - \mathbf{b} \right\|_2^2.
\end{equation}

Solving Eq.~\eqref{eq:opt} exactly is NP-hard for general lattices~\citep{micciancio2002hardness}. A classical and efficient approximation is given by the \emph{Babai point}~\citep{babai1986lovasz}, which performs successive orthogonal projections followed by greedy rounding along a triangular factorization of $\mathbf{A}$. It matches the recent analyses~\citep{chen2025geometryllmquantizationgptq} which suggest that layer-wise solvers such as GPTQ can be interpreted as greedy lattice decoding methods under alternative objective formulations. While the objectives may differ, the underlying principle—approximating a structured lattice decoding problem—remains the same.

% \subsection{Solving BILS via Babai Decoding}
% \label{subsec:babai_quant}
% Define $\bar{\mathbf{A}}=\mathbf{AD}$ then Eq \eqref{eq:opt} becomes
% \begin{equation}
% \label{eq:newILS}
% \min_{\mathbf{q} \in \mathbb{B}^m}
% \left\| \bar{\mathbf{A}}\mathbf{q} - \mathbf{b} \right\|_2^2.
% \end{equation}
% Let the real least squares solution of Eq \eqref{eq:newILS} be denoted by $\bar{\mathbf{q}}$.Then
% \begin{equation}
%     \mathbf{D}\mathbf{A}^\top\mathbf{A}\mathbf{D} \bar{\mathbf{q}} = \mathbf{D}\mathbf{A}^\top\mathbf{b} = \mathbf{D}\mathbf{A}^\top(\mathbf{y}+\mathbf{A}\mathbf{D}\mathbf{z}). 
% \end{equation}
% Then we have 
% \begin{equation}
%     \mathbf{A^\top AD}(\bar{\mathbf{q}}-\mathbf{z})=\mathbf{A}^\top\mathbf{y}
% \end{equation}
% Then we can get $\mathbf{R}$ from the Cholesky factorization $\mathbf{A^\top A}$ Then
% \begin{equation}
%     \mathbf{R^\top R D}(\bar{\mathbf{q}}-\mathbf{z})=\mathbf{A}^\top\mathbf{y}
% \end{equation}
% Set $\mathbf{v}=\mathbf{D}(\bar{\mathbf{q}}-\mathbf{z})$
% Then we can find $\bar{\mathbf{q}}=\mathbf{D}^{-1}\mathbf{v}+\mathbf{z}=\mathbf{v}\oslash\mathbf{s}$where $\oslash$ means component-wise division.
% Denote $\bar{\mathbf{R}}=\mathbf{RD}$ it is easy to verify that Eq \eqref{eq:newILS}
\subsection{Babai-Based Quantization}
\label{subsec:babai_quant}

Define $\bar{\mathbf{A}} := \mathbf{A}\mathbf{D}$. Then Eq.~\eqref{eq:opt} can be rewritten as the box-constrained integer least squares (ILS) problem
\begin{equation}
\label{eq:newILS}
\min_{\mathbf{q} \in \mathbb{B}^m}
\left\| \bar{\mathbf{A}}\mathbf{q} - \mathbf{b} \right\|_2^2 .
\end{equation}

Let $\bar{\mathbf{q}} \in \mathbb{R}^m$ denote the unconstrained real least-squares solution of Eq.~\eqref{eq:newILS}. The corresponding normal equations are given by
\begin{equation*}
\label{eq:normal_eq}
\mathbf{D}\mathbf{A}^\top \mathbf{A}\mathbf{D}\,\bar{\mathbf{q}}
= \mathbf{D}\mathbf{A}^\top \mathbf{b}
= \mathbf{D}\mathbf{A}^\top \bigl( \mathbf{y} + \mathbf{A}\mathbf{D}\mathbf{z} \bigr).
\end{equation*}
Rearranging terms yields
\begin{equation*}
\label{eq:center_shift}
\mathbf{A}^\top \mathbf{A}\mathbf{D}(\bar{\mathbf{q}} - \mathbf{z})
= \mathbf{A}^\top \mathbf{y}.
\end{equation*}

Let $\mathbf{A}^\top \mathbf{A} = \mathbf{R}^\top \mathbf{R}$ be the Cholesky factorization, where $\mathbf{R}$ is upper triangular. Substituting this factorization into Eq.~\eqref{eq:center_shift} gives
\begin{equation*}
\label{eq:chol_eq}
\mathbf{R}^\top \mathbf{R}\mathbf{D}(\bar{\mathbf{q}} - \mathbf{z})
= \mathbf{A}^\top \mathbf{y}.
\end{equation*}

Define the transformed variable
\[
\mathbf{v} := \mathbf{D}(\bar{\mathbf{q}} - \mathbf{z}),
\]
which implies
\[
\bar{\mathbf{q}} = \mathbf{D}^{-1}\mathbf{v} + \mathbf{z}
= \mathbf{v} \oslash \mathbf{s},
\]
where $\oslash$ denotes element-wise division.

Finally, define $\bar{\mathbf{R}} := \mathbf{R}\mathbf{D}$. It is straightforward to verify that Eq.~\eqref{eq:newILS} is equivalent to
\begin{equation}
\label{eq:triangular_ils}
\min_{\mathbf{q} \in \mathbb{B}^m}
\left\| \bar{\mathbf{R}}(\mathbf{q} - \mathbf{z}) - \mathbf{R}^{-\top}\mathbf{A}^\top \mathbf{y} \right\|_2^2,
\end{equation}
which admits efficient approximate solutions via Babai-type successive rounding.
\begin{algorithm}[h] 
\caption{Quantization with \emph{JTA}} \label{a:babai_d}
\begin{algorithmic}[1]
\STATE Compute ${\mathbf{s}}$ and $\mathbf{z}$

\STATE Compute the Cholesky factorization: 
$\widetilde{\mathbf{X}}^\top\widetilde{\mathbf{X}} + \lambda^2\mathbf{I} =\mathbf{R}^\top\mathbf{R}$
\STATE Solve the lower triangular system $\mathbf{R}^\top\mathbf{u}=\mathbf{A}^\top\mathbf{y}$
and the upper triangular system $\mathbf{R}\mathbf{v}=\mathbf{u}$.
\STATE Compute $\bar{\mathbf{q}}=\mathbf{v}\oslash\mathbf{s} +\mathbf{z}$
\STATE Compute $\bar{\mathbf{R}}=\mathbf{RD}$
\STATE $\mathbf{c}(m) = \hat{\mathbf{q}}(m)$  
\STATE $\mathbf{q}(m)= \text{clamp}(\lfloor \mathbf{c}(m) \rceil; \mathbb{B})$  
\FOR{$i=m-1:-1:1$}  
\STATE $\mathbf{c}(i) = \bar{\mathbf{q}}(i) + \left(\sum_{j=i+1}^m \bar{\mathbf{R}}(i,j)(\bar{\mathbf{q}}(j)-\mathbf{q}(j))\right)/\bar{\mathbf{R}}(i,i)$ \\
\STATE $\mathbf{q}(i) = \text{clamp}(\lfloor \mathbf{c}(i) \rceil;\mathbb{B}) $ \\
\ENDFOR
\end{algorithmic}
\end{algorithm}
% ------------------------------------------------------------

\subsection{Random-$K$ Babai/Klein Decoding (Candidate Generation)}
\label{subsec:randomk}
However, as previously mentioned, the Babai algorithm typically produces a suboptimal solution due to its greedy rounding strategy, which does not fully explore the combinatorial structure of the underlying lattice. To address this limitation, Klein \citep{10.5555/338219.338661} proposed a randomized integer least squares (ILS) solver that introduces controlled randomness into the rounding process, enabling the algorithm to explore a richer set of lattice points while maintaining polynomial-time complexity.

\paragraph{Klein-style randomized rounding.}
In each back-substitution step, instead of deterministic rounding such as the line 6 or 10 in Algorithm \ref{a:babai_d}, we sample $\mathbf{q}_i$ from a distribution concentrated around $c_i$: 
\begin{equation}
\label{eq:klein_prob}
    \Pr(\mathbf{q}_i=v)\;=\;
    \frac{\exp\!\left(-\alpha\,\bar{\mathbf{R}}_{ii}\,(\mathbf{c}_i- v)^2\right)}
    {\sum_{u=0}^{Q_{\max}} \exp\!\left(-\alpha\,\bar{\mathbf{R}}_{ii}\,(\mathbf{c}_i- u)^2\right)},
    \quad v\in\mathcal{B},
\end{equation}
where $\alpha>0$ controls the sampling temperature.
Larger $\alpha$ approaches greedy Babai; smaller $\alpha$ encourages exploration.
When $K{=}1$ and $\alpha\to\infty$, the method reduces to deterministic Babai.
To further enhance performance, we are inspired by the \emph{K-best Klein} strategy. Specifically, the randomized Klein algorithm is executed independently for $K$ trials, each producing a candidate integer solution. Among these $K$ candidates, we select the one that minimizes the corresponding residual, yielding a more accurate approximation to the underlying integer least squares objective.See Figure \ref{fig:Residual_over_K} on the performance .
For this paper, we followed \citep{alpha_choice} design choices where $\alpha=ln(\rho)/\min_{1\leq i \leq m}r_{ii}^2$ and $\rho$ is the solution of $K=(e\rho)^{(2m/\rho)}$
 This formulation yields a data-driven trade-off between exploration and exploitation, adaptively adjusting the sampling sharpness according to the lattice geometry and the target list size $K$.
\paragraph{Reference greedy path.}
We reserve a reference path that uses deterministic rounding, guaranteeing inclusion of the Babai solution in the candidate set.

% \begin{algorithm}[t]
% \caption{Random-$K$ Babai/Klein Candidate Generation (Anchor + Sampling)}
% \label{alg:kbest}
% \begin{algorithmic}[1]
% \STATE Initialize best residual $E_{\min}\leftarrow\infty$, best code $q^\star\leftarrow 0$
% \FOR{$t=1$ to $K$}
%     \mathbf{q}^{t}=BKB(\mathbf{R}
%     \STATE $E^{(t)} \leftarrow \|\bar{\mathbf{R}}(\mathbf{q}^{(t)}-\bar{\mathbf{q}})\|_2^2$
%     \IF{$E^{(t)}<E_{\min}$}
%         \STATE $E_{\min}\leftarrow E^{(t)}$, $q^\star\leftarrow \mathbf{q}^{(t)}$
%     \ENDIF
% \ENDFOR
% \STATE \textbf{return} $\mathbf{q}^\star$
% \end{algorithmic}
% \end{algorithm}
\begin{table*}[t]
\small
\begin{tabular}{ccccccccc}
\hline
 &
  Method &
  L2-7B &
  L2-13B &
  L3-8B &
  Q3-0.6B &
  Q3-4B &
  Q3-8B &
  M-7B \\ \hline
\textbf{} &
  BF16 &
  6.97/5.47 &
  6.47/4.88 &
  8.93/6.13 &
  25.28/20.94 &
  16.53/13.67 &
  13.23/9.72 &
  8.15/5.50 \\ \hline
 &
  RTN &
  7.72/6.11 &
  6.83/5.20 &
  12.09/8.53 &
  42.20/37.52 &
  20.15/17.52 &
  15.54/11.97 &
  8.96/6.12 \\
 &
  GPTQ &
  \best{7.11}/5.62 &
  \second{6.56}/4.99 &
  9.41/6.54 &
  30.79/25.20 &
  16.73/\second{13.66} &
  13.51/10.08 &
  8.29/5.63 \\
 &
  AWQ &
  7.16/\textbf{5.61} &
  \second{6.56}/\best{4.97} &
  9.40/6.54 &
  \second{27.99}/25.89 &
  17.18/18.43 &
  \cellcolor[HTML]{FFFFFF}13.51/10.02 &
  8.28/5.61 \\
 &
  \cellcolor[HTML]{C2CFC9}Ours(N) &
  \cellcolor[HTML]{C2CFC9}7.15/5.63 &
  \cellcolor[HTML]{C2CFC9}\second{6.56}/4.99 &
  \cellcolor[HTML]{C2CFC9}9.38/6.52 &
  \cellcolor[HTML]{C2CFC9}28.35/\second{23.48} &
  \cellcolor[HTML]{C2CFC9}18.20/13.97 &
  \cellcolor[HTML]{C2CFC9}13.42/9.97 &
  \cellcolor[HTML]{C2CFC9}\second{8.27}/5.60 \\
 &
  \cellcolor[HTML]{A5B7AF}Ours(R) &
  \cellcolor[HTML]{A5B7AF}\second{7.14}/\second{5.62} &
  \cellcolor[HTML]{A5B7AF}\best{6.55}/\second{4.98} &
  \cellcolor[HTML]{A5B7AF}\second{9.35/6.50} &
  \cellcolor[HTML]{A5B7AF}28.07/23.64 &
  \cellcolor[HTML]{A5B7AF}\second{16.44}/13.84 &
  \cellcolor[HTML]{A5B7AF}\second{13.40/9.95} &
  \cellcolor[HTML]{A5B7AF}\second{8.27}/\second{5.59}  \\
\multirow{-6}{*}{\textbf{\begin{tabular}[c]{@{}c@{}}g128\\ W4A16\end{tabular}}} &
  \cellcolor[HTML]{92A89E}Ours &
  \cellcolor[HTML]{92A89E}\second{7.14}/\best{5.61} &
  \cellcolor[HTML]{92A89E}\best{6.55}/\best{4.97} &
  \cellcolor[HTML]{92A89E}\best{9.33}/\best{6.48} &
  \cellcolor[HTML]{92A89E}\best{26.84/22.80} &
  \cellcolor[HTML]{92A89E}\best{16.32/13.54} &
  \cellcolor[HTML]{92A89E}\best{13.39/9.94} &
  \cellcolor[HTML]{92A89E}\best{8.23/5.58} \\ \hline
 &
  RTN &
  \cellcolor[HTML]{FFFFFF}4e2/5e2 &
  12.50/10.68 &
  4e2/2e3 &
  1e5/2e5 &
  3e3/3e3 &
  5e2/7e2 &
  19.65/14.88 \\
 &
  GPTQ &
  7.92/6.44 &
  6.99/5.43 &
  12.25/8.35 &
  43.60/41.54 &
  18.16/15.19 &
  \second{14.34/11.01} &
  9.10/6.27 \\
 &
  AWQ &
  7.90/\textbf{6.25} &
  \second{6.95}/\best{5.33} &
  11.63/8.19 &
  \second{39.60/36.88} &
  19.75/16.83 &
  14.97/11.23 &
  8.89/6.07 \\
 &
  \cellcolor[HTML]{C2CFC9}Ours(N) &
  \cellcolor[HTML]{C2CFC9}7.88/6.33 &
  \cellcolor[HTML]{C2CFC9}6.98/5.41 &
  \cellcolor[HTML]{C2CFC9}11.52/8.19 &
  \cellcolor[HTML]{C2CFC9}47.23/42.75 &
  \cellcolor[HTML]{C2CFC9}25.46/20.36 &
  \cellcolor[HTML]{C2CFC9}14.44/11.11 &
  \cellcolor[HTML]{C2CFC9}8.87/\second{6.06} \\
 &
  \cellcolor[HTML]{A5B7AF}Ours(R) &
  \cellcolor[HTML]{A5B7AF}\second{7.87}/6.34 &
  \cellcolor[HTML]{A5B7AF}\best{6.94}/\second{5.39} &
  \cellcolor[HTML]{A5B7AF}\second{11.43/8.16} &
  \cellcolor[HTML]{A5B7AF}44.09/40.19 &
  \cellcolor[HTML]{A5B7AF}\second{19.52/14.91} &
  \cellcolor[HTML]{A5B7AF}14.37/11.05 &
  \cellcolor[HTML]{A5B7AF}\second{8.86}/\best{6.03} \\
\multirow{-6}{*}{\textbf{\begin{tabular}[c]{@{}c@{}}g128\\ W3A16\end{tabular}}} &
  \cellcolor[HTML]{92A89E}Ours &
  \cellcolor[HTML]{92A89E}\best{7.86}/\second{6.30} &
  \cellcolor[HTML]{92A89E}\best{6.94}/\best{5.33} &
  \cellcolor[HTML]{92A89E}\best{11.40}/\best{7.87} &
  \cellcolor[HTML]{92A89E}\best{35.03/31.75} &
  \cellcolor[HTML]{92A89E}\best{17.01/12.89} &
  \cellcolor[HTML]{92A89E}\best{14.30/10.98} &
  \cellcolor[HTML]{92A89E}\best{8.84/6.03} \\ \hline
 &
  GPTQ &
  7.37/5.83 &
  6.70/5.16 &
  10.30/7.37 &
  37.03/33.64 &
  17.34/14.34 &
  13.99/10.65 &
  8.53/5.81 \\
 &
  AWQ &
  7.34/5.83 &
  6.68/5.06 &
  9.44/7.10 &
  30.07/25.89 &
  18.83/16.35 &
  14.29/10.77 &
  10.24/5.88 \\
 &
  QUIP &
  12.54/10.44 &
  6.71/5.15 &
  9.84/6.85 &
  51.41/47.10 &
  182.16/126.55 &
  19.19/14.85 &
  9.69/6.65 \\
 &
  \cellcolor[HTML]{C2CFC9}Ours(N) &
  \cellcolor[HTML]{C2CFC9}7.14/5.64 &
  \cellcolor[HTML]{C2CFC9}\second{6.57}/4.98 &
  \cellcolor[HTML]{C2CFC9}9.39/6.52 &
  \cellcolor[HTML]{C2CFC9}28.35/23.48 &
  \cellcolor[HTML]{C2CFC9}\second{17.30/14.30}&
  \cellcolor[HTML]{C2CFC9}\second{13.41}/9.97 &
  \cellcolor[HTML]{C2CFC9}8.28/5.62 \\
 &
  \cellcolor[HTML]{A5B7AF}Ours(R) &
  \cellcolor[HTML]{A5B7AF}\second{7.11}/\second{5.62} &
  \cellcolor[HTML]{A5B7AF}\best{6.56}/\second{4.97} &
  \cellcolor[HTML]{A5B7AF}\second{9.38/6.50} &
  \cellcolor[HTML]{A5B7AF}\second{28.11/23.04} &
  \cellcolor[HTML]{A5B7AF}\second{17.30/14.30} &
  \cellcolor[HTML]{A5B7AF}\second{13.41/9.96} &
  \cellcolor[HTML]{A5B7AF}\second{8.27/5.60} \\
\multirow{-6}{*}{\textbf{W4A16}} &
  \cellcolor[HTML]{92A89E}Ours &
  \cellcolor[HTML]{92A89E}\best{7.10}/\best{5.60} &
  \cellcolor[HTML]{92A89E}\best{6.56}/\best{4.96} &
  \cellcolor[HTML]{92A89E}\best{9.33/6.48} &
  \cellcolor[HTML]{92A89E}\best{26.94/22.61} &
  \cellcolor[HTML]{92A89E}\best{17.25/14.27} &
  \cellcolor[HTML]{92A89E} \best{13.38/9.92}&
  \cellcolor[HTML]{92A89E}\best{8.24/5.59} \\ \hline
 &
  GPTQ &
  9.83/8.33 &
  8.04/6.52 &
  29.42/23.81 &
  90.23/90.98 &
  23.16/21.39 &
  18.10/16.46 &
  10.90/8.01 \\
 &
  AWQ &
  15.62/15.51 &
  8.17/6.45 &
  17.60/11.84 &
  \cellcolor[HTML]{FFFFFF}78.52/79.44 &
  30.31/33.19 &
  18.36/\second{14.97} &
  10.27/7.65 \\
 &
  QUIP &
  27.47/28.05 &
  7.17/5.57 &
  11.50/8.31 &
  121.26/112.09 &
  23.02/19.74 &
  30.39/19.78 &
  9.30/6.61 \\
 &
  \cellcolor[HTML]{C2CFC9}Ours(N) &
  \cellcolor[HTML]{C2CFC9}7.88/6.30 &
  \cellcolor[HTML]{C2CFC9}\second{6.97}/5.42 &
  \cellcolor[HTML]{C2CFC9}11.52/8.17 &
  \cellcolor[HTML]{C2CFC9}47.23/42.75 &
  \cellcolor[HTML]{C2CFC9}22.95/19.65&
  \cellcolor[HTML]{C2CFC9}25.46/20.36 &
  \cellcolor[HTML]{C2CFC9}8.78/6.09 \\
 &
  \cellcolor[HTML]{A5B7AF}Ours(R) &
  \cellcolor[HTML]{A5B7AF}\second{7.87}/\second{6.29} &
  \cellcolor[HTML]{A5B7AF}\second{6.97}/\second{5.41} &
  \cellcolor[HTML]{A5B7AF}\second{11.35/8.12} &
  \cellcolor[HTML]{A5B7AF}\second{43.94/39.77} &
  \cellcolor[HTML]{A5B7AF}\second{22.90/19.63} &
  \cellcolor[HTML]{A5B7AF}\second{18.07/}16.48 &
  \cellcolor[HTML]{A5B7AF}\second{8.77/6.07} \\
\multirow{-6}{*}{\textbf{W3A16}} &
  \cellcolor[HTML]{92A89E}Ours &
  \cellcolor[HTML]{92A89E}\best{7.75/6.22} &
  \cellcolor[HTML]{92A89E}\best{6.96}/\best{5.40} &
  \cellcolor[HTML]{92A89E}\best{11.38/8.04} &
  \cellcolor[HTML]{92A89E}\best{35.56/32.95} &
  \cellcolor[HTML]{92A89E}\best{22.87/19.56}&
  \cellcolor[HTML]{92A89E} \best{18.05/14.75}&
  \cellcolor[HTML]{92A89E}\best{8.72/6.06} \\ \cline{1-9} 
\end{tabular}
\caption{Perplexity comparison across different models and quantization methods. For each entry, the left value is evaluated on C4 and the right value on WikiText-2. We denote our proposed methods as follows: \textbf{Ours(N)} represents Naïve Babai, \textbf{Ours(R)} represents Random-$K$ Babai, and \textbf{Ours} represents Random-$K$ Babai with Joint Activation-Target optimization. The model families are abbreviated as: L2/L3 for Llama2/Llama 3, Q3 for the Qwen3 series, and M for Mistral. All activations are in BF16 format.}
\label{tab:ppl}
\end{table*}
\subsubsection{GPU-efficient Path-Isolated Random-$K$ Babai Search}
\label{subsec:ppi_kbabai}

Naively parallelizing $K$ decoding paths on GPU can introduce \emph{cross-path interference} when residual states are shared or updated in-place after paths diverge, leading to incorrect centers $c_i$ and biased sampling.
We introduce a GPU-efficient random $K$-best Babai solver with strict path isolation, termed
\textbf{Parallel Path-Isolated $K$-best Babai (PPI-KBabai)} (or \textbf{Path-Isolated Random-$K$ Babai Search}).
Unlike naive parallel $K$-path implementations that share residual states, our method maintains an independent residual buffer per candidate path and performs blocked look-ahead updates via batched matrix multiplication.
We further reserve an reference greedy path to guarantee inclusion of the deterministic Babai solution, and select the best candidate using the unified JTA score in Sec.~\ref{subsec:jta}.
The Klein Styled JTA quantization, K-Best JTA quantiation and detailed kernel design, buffer layout, and blocked update rules are provided in Appendix~\ref{app}.

%% file: sections/4_experiment.tex
\definecolor{BFgray}{gray}{0.90}
\section{Experiments}

To evaluate the effectiveness of our method, we conduct comprehensive experiments comparing it against commonly used layer-wise quantization baselines. We first describe the baseline methods, the large language models being quantized, as well as the evaluation metrics and datasets. We then present the experimental results, followed by a detailed analysis. To ensure fair comparisons, we adopt the default configurations of each baseline's repo whenever possible.
\paragraph{Baselines}

\begin{figure}
    \centering
    \includegraphics[width=1\linewidth]{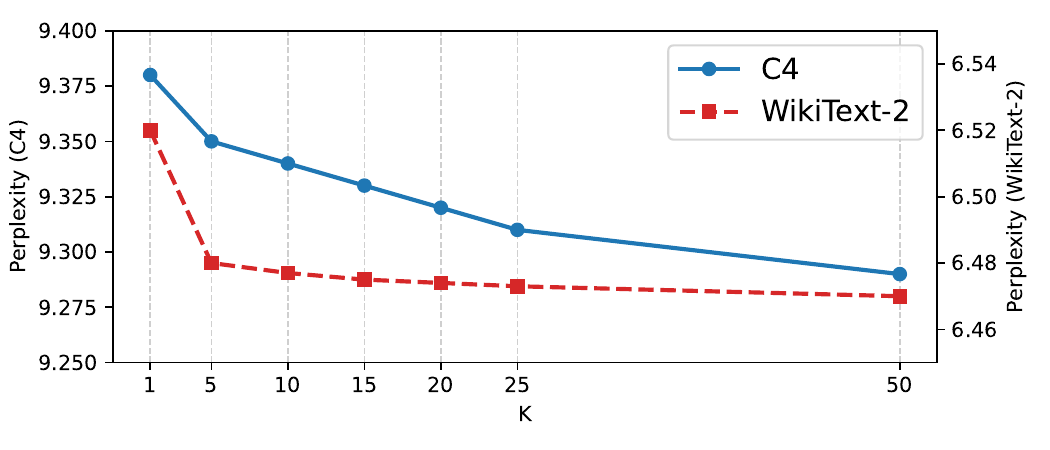}
    \caption{\textbf{Ablation study on the candidate size $K$.} We evaluate the perplexity on C4 and WikiText-2 datasets using Llama-3-8B with 4-bit quantization (group size 128). The results demonstrate the impact of increasing the search space $K$.}
    \label{fig:ppl_over_K}
\end{figure}

We evaluate our method against state-of-the-art quantization approaches, including AWQ \citep{awq}, GPTQ \citep{chen2023frugalgpt}, and QUIP \citep{chee2024quip2bitquantizationlarge}, and additionally report round-to-nearest as a naïve baseline for perplexity.Our method is evaluated under three settings: Naïve Babai (Ours-N), Random-$K$ Babai (Ours-R), and Random $K$ Babai with Joint Activation-Target optimization (Ours).Although theortically, the $K$ can be any positive integer number, considering the trade-off between resource usage and performance gain Figure\ref{fig:ppl_over_K}, we set the $K$ as 5 in our experiments.  To ensure strong baselines, we enable activation ordering for GPTQ. Calibration is performed using 128 samples from the C4 \citep{c4} dataset with a sequence length of 2048 tokens. For all baselines, only minimal modifications are made to accommodate the architectural structures of newer LLMs, while their original algorithms and configurations are kept unchanged.
\begin{table*}[t]
\small
\setlength{\tabcolsep}{3pt}
\renewcommand{\arraystretch}{1.05}
\centering
\resizebox{1.0\linewidth}{!}{%
\begin{tabular}{cc|ccccccc|ccccccccccccc}
\toprule
\multirow{2}{*}{Model} & \multirow{2}{*}{Method} &\multicolumn{7}{c}{4 bits} & \multicolumn{7}{c}{3 bits} \\
\cmidrule(lr){3-9} \cmidrule(lr){10-16}
 &  & ARC-C & ARC-E & BoolQ & Hella & PIQA & Wino & Average* & ARC-C & ARC-E & BoolQ & Hella & PIQA & Wino & Average* \\
\midrule
\midrule
\multirow{7}{*}{\rotatebox[origin=c]{90}{Llama3-8B}}
& BF16 & \cellcolor{BFgray}{51.71} & \cellcolor{BFgray}{80.89} & \cellcolor{BFgray}{81.77} & \cellcolor{BFgray}{60.56} & \cellcolor{BFgray}{78.84} & \cellcolor{BFgray}{73.40} & \cellcolor{BFgray}{71.20}
& \cellcolor{BFgray}{51.71} & \cellcolor{BFgray}{80.89} & \cellcolor{BFgray}{81.77} & \cellcolor{BFgray}{60.56} & \cellcolor{BFgray}{78.84} & \cellcolor{BFgray}{73.40} & \cellcolor{BFgray}{71.20} \\
& GPTQ & 48.29 & 78.89 & 80.21 & 59.47 & 78.29 & \best{73.32} & 69.74
& 34.21 & 66.04 & 71.89 & 55.26 & 73.56 & 68.11 & 61.51 \\
& AWQ & 48.70 & \second{79.21} & 80.80 & 59.73 & 78.62 & \best{73.32} & 70.06
& \best{43.86} & 75.21 & 76.15 & 55.79 & 75.95 & 71.03 & 66.33 \\
& QUIP & 46.25 & 77.36 & \best{81.93} & 58.76 & 77.53 & 73.09 & 69.15
& 41.38 & 71.34 & 77.68 & 55.67 & 76.12 & \best{71.67} & 65.64 \\
& O(N) & \cellcolor[HTML]{C2CFC9}48.46 & \cellcolor[HTML]{C2CFC9}78.53 & \cellcolor[HTML]{C2CFC9}81.04 & \cellcolor[HTML]{C2CFC9}\best{60.00} & \cellcolor[HTML]{C2CFC9}78.18 & \cellcolor[HTML]{C2CFC9}73.16 & \cellcolor[HTML]{C2CFC9}69.90
& \cellcolor[HTML]{C2CFC9}41.81 & \cellcolor[HTML]{C2CFC9}71.71 & \cellcolor[HTML]{C2CFC9}76.73 & \cellcolor[HTML]{C2CFC9}\best{56.71} & \cellcolor[HTML]{C2CFC9}76.88 & \cellcolor[HTML]{C2CFC9}71.35 & \cellcolor[HTML]{C2CFC9}65.87 \\
& O(R) & \cellcolor[HTML]{A5B7AF}\best{49.32} & \cellcolor[HTML]{A5B7AF}79.12 & \cellcolor[HTML]{A5B7AF}81.04 & \cellcolor[HTML]{A5B7AF}\second{59.79} & \cellcolor[HTML]{A5B7AF}\second{78.24} & \cellcolor[HTML]{A5B7AF}72.30 & \cellcolor[HTML]{A5B7AF}\second{69.97}
& \cellcolor[HTML]{A5B7AF}41.30 & \cellcolor[HTML]{A5B7AF}\second{73.48} & \cellcolor[HTML]{A5B7AF}76.45 & \cellcolor[HTML]{A5B7AF}\second{56.67} & \cellcolor[HTML]{A5B7AF}\second{77.37} & \cellcolor[HTML]{A5B7AF}70.64 & \cellcolor[HTML]{A5B7AF}\second{65.99} \\
& O & \cellcolor[HTML]{92A89E}\second{49.15} & \cellcolor[HTML]{92A89E}\best{79.59} & \cellcolor[HTML]{92A89E}\second{81.13} & \cellcolor[HTML]{92A89E}59.70 & \cellcolor[HTML]{92A89E}\best{78.67} & \cellcolor[HTML]{92A89E}\second{72.69} & \cellcolor[HTML]{92A89E}\best{70.16}
& \cellcolor[HTML]{92A89E}\second{42.92} & \cellcolor[HTML]{92A89E}\best{75.42} & \cellcolor[HTML]{92A89E}\best{78.69} & \cellcolor[HTML]{92A89E}55.73 & \cellcolor[HTML]{92A89E}\best{77.75} & \cellcolor[HTML]{92A89E}\second{70.32} & \cellcolor[HTML]{92A89E}\best{66.81} \\
\midrule
\multirow{7}{*}{\rotatebox[origin=c]{90}{Qwen3-4B}}
& BF16 & \cellcolor{BFgray}{50.34} & \cellcolor{BFgray}{80.26} & \cellcolor{BFgray}{85.11} & \cellcolor{BFgray}{52.29} & \cellcolor{BFgray}{75.30} & \cellcolor{BFgray}{65.67} & \cellcolor{BFgray}{68.16}
& \cellcolor{BFgray}{50.34} & \cellcolor{BFgray}{80.26} & \cellcolor{BFgray}{85.11} & \cellcolor{BFgray}{52.29} & \cellcolor{BFgray}{75.30} & \cellcolor{BFgray}{65.67} & \cellcolor{BFgray}{68.16} \\
& GPTQ & 47.57 & 77.90 & 82.43 & 50.12 & \best{74.43} & 62.43 & 65.81
& 39.33 & 70.92 & \best{80.95} & \best{46.81} & 70.92 & \best{61.64} & 61.76 \\
& AWQ & 46.29 & 77.32 & 83.23 & 50.37 & \second{74.10} & \best{62.88} & 65.70
& 40.27 & 68.27 & 81.19 & 46.02 & 70.87 & 60.38 & 61.17 \\
& QUIP & 39.25 & 71.09 & 80.73 & 43.64 & 70.24 & 59.91 & 60.81
& 24.15 & 47.81 & 67.86 & 33.72 & 62.79 & 51.14 & 47.91 \\
& O(N) & \cellcolor[HTML]{C2CFC9}41.81 & \cellcolor[HTML]{C2CFC9}74.79 & \cellcolor[HTML]{C2CFC9}\second{83.46} & \cellcolor[HTML]{C2CFC9}47.61 & \cellcolor[HTML]{C2CFC9}73.12 & \cellcolor[HTML]{C2CFC9}58.33 & \cellcolor[HTML]{C2CFC9}63.19
& \cellcolor[HTML]{C2CFC9}35.95 & \cellcolor[HTML]{C2CFC9}58.68 & \cellcolor[HTML]{C2CFC9}69.20 & \cellcolor[HTML]{C2CFC9}40.65 & \cellcolor[HTML]{C2CFC9}67.74 & \cellcolor[HTML]{C2CFC9}50.83 & \cellcolor[HTML]{C2CFC9}53.84 \\
& O(R) & \cellcolor[HTML]{A5B7AF}44.28 & \cellcolor[HTML]{A5B7AF}74.03 & \cellcolor[HTML]{A5B7AF}82.97 & \cellcolor[HTML]{A5B7AF}\second{49.28} & \cellcolor[HTML]{A5B7AF}72.80 & \cellcolor[HTML]{A5B7AF}\second{61.17} & \cellcolor[HTML]{A5B7AF}\second{64.09}
& \cellcolor[HTML]{A5B7AF}\second{37.79} & \cellcolor[HTML]{A5B7AF}\second{65.35} & \cellcolor[HTML]{A5B7AF}\second{70.58} & \cellcolor[HTML]{A5B7AF}\second{43.39} & \cellcolor[HTML]{A5B7AF}67.25 & \cellcolor[HTML]{A5B7AF}\second{53.99} & \cellcolor[HTML]{A5B7AF}\second{56.39} \\
& O & \cellcolor[HTML]{92A89E}\best{46.65} & \cellcolor[HTML]{92A89E}\best{78.27} & \cellcolor[HTML]{92A89E}\best{84.31} & \cellcolor[HTML]{92A89E}\best{49.85} & \cellcolor[HTML]{92A89E}74.21 & \cellcolor[HTML]{92A89E}62.35 & \cellcolor[HTML]{92A89E}\best{65.94}
& \cellcolor[HTML]{92A89E}\best{41.50} & \cellcolor[HTML]{92A89E}\best{71.41} & \cellcolor[HTML]{92A89E}80.95 & \cellcolor[HTML]{92A89E}45.61 & \cellcolor[HTML]{92A89E}\best{72.64} & \cellcolor[HTML]{92A89E}60.06 & \cellcolor[HTML]{92A89E}\best{62.03} \\
\midrule
\multirow{7}{*}{\rotatebox[origin=c]{90}{Qwen3-8B}}
& BF16 & \cellcolor{BFgray}{55.38} & \cellcolor{BFgray}{83.50} & \cellcolor{BFgray}{86.73} & \cellcolor{BFgray}{57.12} & \cellcolor{BFgray}{76.50} & \cellcolor{BFgray}{67.88} & \cellcolor{BFgray}{71.19}
& \cellcolor{BFgray}{55.38} & \cellcolor{BFgray}{83.50} & \cellcolor{BFgray}{86.73} & \cellcolor{BFgray}{57.12} & \cellcolor{BFgray}{76.50} & \cellcolor{BFgray}{67.88} & \cellcolor{BFgray}{71.19} \\
& GPTQ & \best{54.27} & \best{82.74} & \best{86.73} & \best{56.62} & \second{76.22} & 66.77 & 70.56
& 44.28 & 74.62 & 83.33 & 53.88 & 75.95 & 67.56 & 66.60 \\
& AWQ & \second{53.67} & \second{82.37} & \second{85.93} & \second{55.91} & 75.79 & \best{68.75} & \second{70.40}
& \best{50.56} & \best{80.93} & \best{84.43} & 52.80 & 75.90 & 67.25 & \best{68.65} \\
& QUIP & 45.48 & 75.42 & 81.99 & 44.72 & 69.53 & 62.67 & 63.30
& 40.27 & 71.93 & 77.37 & 43.22 & 71.33 & 59.83 & 60.66 \\
& O(N) & \cellcolor[HTML]{C2CFC9}53.75 & \cellcolor[HTML]{C2CFC9}82.20 & \cellcolor[HTML]{C2CFC9}86.02 & \cellcolor[HTML]{C2CFC9}56.39 & \cellcolor[HTML]{C2CFC9}\second{76.22} & \cellcolor[HTML]{C2CFC9}68.03 & \cellcolor[HTML]{C2CFC9}70.44
& \cellcolor[HTML]{C2CFC9}\second{47.69} & \cellcolor[HTML]{C2CFC9}77.61 & \cellcolor[HTML]{C2CFC9}84.46 & \cellcolor[HTML]{C2CFC9}53.76 & \cellcolor[HTML]{C2CFC9}74.76 & \cellcolor[HTML]{C2CFC9}67.35 & \cellcolor[HTML]{C2CFC9}67.61 \\
& O(R) & \cellcolor[HTML]{A5B7AF}53.41 & \cellcolor[HTML]{A5B7AF}82.36 & \cellcolor[HTML]{A5B7AF}86.02 & \cellcolor[HTML]{A5B7AF}56.34 & \cellcolor[HTML]{A5B7AF}76.06 & \cellcolor[HTML]{A5B7AF}\second{68.82} & \cellcolor[HTML]{A5B7AF}70.50
& \cellcolor[HTML]{A5B7AF}47.59 & \cellcolor[HTML]{A5B7AF}\second{77.62} & \cellcolor[HTML]{A5B7AF}84.04 & \cellcolor[HTML]{A5B7AF}\second{53.88} & \cellcolor[HTML]{A5B7AF}74.59 & \cellcolor[HTML]{A5B7AF}\second{67.46} & \cellcolor[HTML]{A5B7AF}67.53 \\
& O & \cellcolor[HTML]{92A89E}54.18 & \cellcolor[HTML]{92A89E}82.62 & \cellcolor[HTML]{92A89E}86.45 & \cellcolor[HTML]{92A89E}55.70 & \cellcolor[HTML]{92A89E}\best{76.39} & \cellcolor[HTML]{92A89E}68.59 & \cellcolor[HTML]{92A89E}\best{70.66}
& \cellcolor[HTML]{92A89E}47.27 & \cellcolor[HTML]{92A89E}78.41 & \cellcolor[HTML]{92A89E}\second{84.39} & \cellcolor[HTML]{92A89E}\best{54.23} & \cellcolor[HTML]{92A89E}\best{76.27} & \cellcolor[HTML]{92A89E}\best{67.90} & \cellcolor[HTML]{92A89E}\second{68.08} \\
\bottomrule
\end{tabular}}
\caption{\textbf{Zero-shot accuracy comparison on six common sense reasoning tasks across Llama3-8B, Qwen3-4B, and Qwen3-8B.} The evaluation is conducted under \textbf{4-bit} and \textbf{3-bit} quantization settings. We denote our proposed methods as: \textbf{O(N)} for Naïve Babai, \textbf{O(R)} for Random $K$-Best Babai, and \textbf{O} for Random $K$-Best Babai with Joint Activation-Target optimization. The unquantized \textbf{BF16} baseline is shaded in gray. For quantized methods, \textbf{bold} indicates the best result and \underline{underlining} denotes the second-best.}
\label{tab:zero-shot}
\end{table*}
\begin{table*}[t]
\centering
\small
\setlength{\tabcolsep}{3.5pt} 
\begin{tabular}{ll cccc c cccc c cccc}
\toprule
 &  & \multicolumn{4}{c}{\textbf{LLaMA3-8B}} & & \multicolumn{4}{c}{\textbf{Qwen3-4B}} & & \multicolumn{4}{c}{\textbf{Qwen3-8B}} \\
\cmidrule(lr){3-6} \cmidrule(lr){8-11} \cmidrule(lr){13-16}
Bit & Method & GSM8K & GPQA & MBPP & Avg & & GSM8K & GPQA & MBPP & Avg & & GSM8K & GPQA & MBPP & Avg \\
\midrule
BF16 & BF16 & 51.86 & 35.98 & 48.40 & 45.41 & & 83.70 & 38.38 & 62.40 & 61.49 & & 88.48 & 33.33 & 65.00 & 62.27 \\
\midrule
 & GPTQ & \underline{45.03} & 29.27 & \underline{45.00} & 39.77 & & 57.71 & \underline{31.81} & 27.60 & 39.77 & & \underline{86.80} & \textbf{36.86} & 60.20 & \underline{61.29} \\
 & AWQ & 44.42 & \underline{32.32} & \textbf{46.40} & \underline{41.05} & & \underline{80.15} & 33.33 & \textbf{57.80} & \underline{57.09} & & 85.44 & 32.83 & \textbf{62.80} & 60.36 \\
 & QUIP & 42.30 & 29.87 & 40.40 & 37.52 & & 5.69 & 25.75 & 0.00 & 10.48 & & 64.40 & 31.31 & 0.00 & 31.90 \\
\multirow{-4}{*}{\shortstack{4-bit\\(g128)}} & \cellcolor[HTML]{92A89E}Ours & 
\cellcolor[HTML]{92A89E}\textbf{48.22} & \cellcolor[HTML]{92A89E}\textbf{32.93} & \cellcolor[HTML]{92A89E}44.80 & \cellcolor[HTML]{92A89E}\textbf{41.98} & & 
\cellcolor[HTML]{92A89E}\textbf{84.29} & \cellcolor[HTML]{92A89E}\textbf{36.36} & \cellcolor[HTML]{92A89E}\underline{56.80} & \cellcolor[HTML]{92A89E}\textbf{59.15} & & 
\cellcolor[HTML]{92A89E}\textbf{87.87} & \cellcolor[HTML]{92A89E}\underline{34.34} & \cellcolor[HTML]{92A89E}\underline{63.20} & \cellcolor[HTML]{92A89E}\textbf{61.80} \\
\bottomrule
\end{tabular}
\caption{\textbf{Reasoning accuracy (\%) on GSM8K, GPQA, and MBPP benchmarks.} We evaluate LLaMA3-8B, Qwen3-4B, and Qwen3-8B. All quantized methods are evaluated with \textbf{4-bit} weights and a group size of \textbf{128 (g128)}. Our method (denoted as \textbf{Ours}) employs \textbf{Random $K$ Babai with Joint Target Activation (JTA) optimization}. The unquantized baseline is reported in \textbf{BF16}. \textbf{Bold} indicates the best result and \underline{underlining} denotes the second-best.}
\label{tab:reasoning}
\end{table*}
\paragraph{Models}
We consider two major LLM families, Qwen \citep{yang2025qwen3technicalreport}and LLaMA \citep{touvron2023llama,grattafiori2024llama3herdmodels}, together with Mistral-7B \citep{jiang2023mistral7b}. This selection covers diverse model families, reasoning capabilities, and parameter sizes.
\paragraph{Metrics and Datasets}
To test the basic text generation ability, we use the perplexity to measure each quantized model on C4 \citep{c4} and Wikitext2  \citep{wiki}. Then we further evaluate on the zero shot accuary on ARC \citep{arc},boolq \citep{boolq},HellaSwag \citep{zellers2019hellaswag},PIQA \citep{bisk2020piqa} and WinoGrande  \citep{sakaguchi2021winogrande} For the last part, we test the reasoning ablility on GSM-8K \citep{GSM}, GPQA-diamond \citep{rein2023gpqagraduatelevelgoogleproofqa} and MBPP \citep{mbpp}. Except the evalution of PPL, all other tasks are evaluated on LM-harness library \citep{eval-harness}
\paragraph{Perplexity} Table \ref{tab:ppl} reports perplexity (PPL) across multiple LLaMA-2/3, Qwen-3, and Mistral models under group-wise weight-only quantization with activation precision fixed at 16-bit. At 4-bit, existing PTQ methods such as GPTQ and AWQ achieve reasonable performance on large LLaMA models but exhibit noticeable degradation or instability on harder architectures, particularly Qwen-0.6B and llama-3-8B, while RTN frequently diverges. In contrast, our method consistently attains the lowest or near-lowest perplexity across all model families, with clear improvements in challenging regimes. The advantage becomes more pronounced at 3-bit, where prior methods often suffer sharp performance drops or catastrophic failures, whereas our approach remains stable and shows smooth degradation relative to BP16. These results indicate that jointly optimizing quantization objectives and search strategies leads to substantially improved robustness under low-bit quantization, especially for smaller and more sensitive models.When group quantization is disabled (group size = 0), baseline methods such as GPTQ, AWQ, and QUIP exhibit a clear increase in quantization error, with substantial performance degradation and, in some cases, numerical instability—particularly for smaller and more sensitive models. In contrast, our method remains stable in this setting and consistently preserves low perplexity across all architectures.

\paragraph{Zero-shot accuracy}
Table \ref{tab:zero-shot} compares BF16 and low-bit PTQ methods across three model families under both 4-bit and 3-bit settings. At 4 bits, all methods remain relatively close to BF16, but our approach consistently achieves the strongest or second-strongest performance across most tasks and models, yielding the highest average accuracy for all three backbones. Notably, the gains are more pronounced on reasoning-heavy benchmarks such as ARC-C, ARC-E, and Hella, indicating improved robustness beyond simple lexical or pattern-matching tasks. When moving to the more challenging 3-bit regime, performance gaps widen substantially: GPTQ and QUIP exhibit significant degradation, particularly on ARC and Hella, while our method degrades more gracefully and maintains clear advantages in average performance. This trend is consistent across model scales, from Qwen-3-4B to Qwen-3-8B, suggesting that the proposed approach scales favorably and is less sensitive to aggressive quantization. Overall, the results demonstrate that our method not only preserves accuracy under moderate compression but also provides superior stability in extreme low-bit settings, highlighting its effectiveness as a general and robust PTQ solution.
\paragraph{Reasoning}
Table \ref{tab:reasoning} reports reasoning accuracy on GSM8K, GPQA, and MBPP under 4-bit quantization across three model families. Compared to GPTQ, AWQ, and QUIP, our method consistently achieves the highest average accuracy for all backbones, with particularly strong gains on GSM8K and GPQA, which require multi-step numerical and factual reasoning. While AWQ remains competitive on MBPP for some models, its performance is less consistent across tasks, leading to lower overall averages. QUIP exhibits severe degradation on reasoning benchmarks, including complete failure on MBPP for Qwen models, highlighting its instability under reasoning-intensive workloads. In contrast, our approach maintains balanced performance across arithmetic, knowledge-intensive, and program synthesis tasks, closely tracking BF16 accuracy despite aggressive quantization. These results indicate that the proposed method better preserves structured reasoning capabilities under low-bit constraints, rather than overfitting to a single task type or model family.

\begin{figure}
    \centering
    \includegraphics[width=1\linewidth]{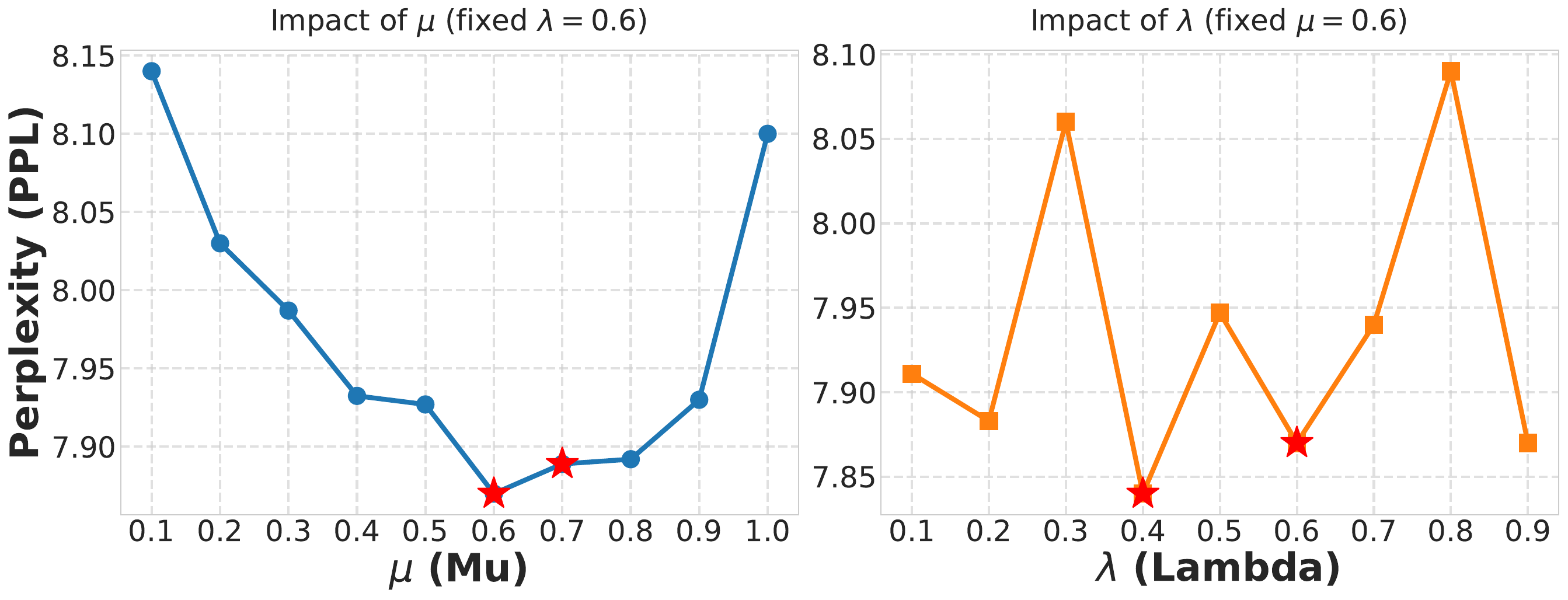}
    \caption{\textbf{Sensitivity analysis of hyperparameters $\mu$ and $\lambda$.} Evaluated on WikiText-2 (calibrated on C4), the plots show the perplexity trend when varying one parameter while fixing the other at 0.6. The U-shaped curve for $\mu$ (left) confirms the necessity of balancing the two objectives, while $\lambda$ (right) shows 0.6 as a robust operating point.}
\label{fig:sensitivity}
    \label{fig:ppl_over_mu_lambda}
\end{figure}

\paragraph{Ablations}
We investigate the impact of the hyperparameter $K$ by varying it within $\{1, 5, \dots, 50\}$ under the setting of 4-bit Llama-3-8B with a group size of 128.
As illustrated in Fig.~\ref{fig:ppl_over_K}, the perplexity on both WikiText-2 and C4 exhibits a \textbf{significant drop} at $K=5$.
Beyond this point, while the performance continues to improve up to $K=50$, the marginal gains become negligible.
Based on these observations, we select $K=5$ as the default configuration for our method.

We analyze the impact of $\mu$ and $\lambda$ on WikiText-2 in Fig.~\ref{fig:ppl_over_mu_lambda}.
For $\mu$, performance is optimized at $0.6$ but degrades at the boundaries ($0.1, 1.0$).
This confirms that neither Eq.~\ref{eq:obj_rt} nor Eq.~\ref{eq:obj_mismatch} alone is sufficient; a balanced combination is essential.
For $\lambda$, despite higher variance, setting $\lambda=0.6$ yields a robust minimum.
Consequently, we adopt $(\mu=0.6, \lambda=0.6)$ as the default configuration for 3 bits setting. Similarly,  we adopt $(\mu=0.1, \lambda=0.2)$ for 4 bits setting.

%% file: sections/5_conclusion.tex
\section{Conclusion}
We presented a unified post-training quantization framework that reformulates layer-wise weight quantization as a structured integer optimization problem and solves it using a joint objective with Babai-type decoding. By integrating compensation-aware targets and controlled randomness, the proposed method consistently outperforms existing PTQ approaches across a wide range of models, tasks, and bit-widths. Extensive experiments on both general evaluation benchmarks and reasoning-intensive tasks demonstrate that our approach preserves accuracy more effectively under aggressive low-bit quantization, while remaining stable across model scales. These results suggest that principled optimization-based formulations provide a robust foundation for future low-bit quantization methods, bridging classical lattice decoding techniques and modern large language model compression.

\paragraph{Limitations}
Our current framework does not yet incorporate weight permutation (e.g., GPTQ) or dynamic scaling, which are promising for future integration.
Additionally, we presently use fixed hyperparameters ($\mu, \lambda$) and candidate size $K$ for all layers.
Future work will explore layer-wise adaptive strategies, specifically assigning distinct $\mu, \lambda$ and varying the number of random candidates per layer to better match the sensitivity of different model components.

%% file: sections/x_appendix.tex
\section{Appendix}
\label{app}
\begin{algorithm}[h]
\caption{Parallel $K$-Path Computation via Vectorized Back-Substitution}
\label{alg:parallel_babai}
\begin{algorithmic}[1]
\REQUIRE Cholesky factor $\mathbf{R} \in \mathbb{R}^{m \times m}$, Continuous weights $\bar{\mathbf{W}} \in \mathbb{R}^{m \times n}$
\REQUIRE Hyperparameters: $K$ (candidates), $B$ (block size)
\ENSURE Candidate Integer Weights $\mathbf{Q} \in \mathbb{Z}^{(K) \times m \times n}$

\STATE \textbf{Initialization:}
\STATE Initialize candidate centers $\mathbf{C} \in \mathbb{R}^{(K) \times m \times n}$ by broadcasting $\bar{\mathbf{W}}$
\STATE Initialize $\mathbf{Q}$ (storage for quantized values)

\COMMENT{\textit{Iterate backwards in blocks for matrix-level efficiency}}
\FOR{row block index $j_{start} = m$ \textbf{down to} $1$ \textbf{step} $B$}
    \STATE Define current block $J = [j_{start}-B : j_{start}]$ and future block $F = [j_{start} : m]$

    \STATE \textbf{1. Global Vectorized Update (Matrix-Matrix Op)}
    \IF{$F$ is not empty}
        \STATE Compute quantization error from processed rows: $\Delta_F = \mathbf{C}_{:, F, :} - \mathbf{Q}_{:, F, :}$
        \STATE Propagate error to all $K$ paths simultaneously using matrix multiplication:
        \STATE $\mathbf{C}_{:, J, :} \leftarrow \mathbf{C}_{:, J, :} + \frac{1}{\text{diag}(\mathbf{R})_J} \left( \mathbf{R}_{J, F} \cdot \Delta_F \right)$
    \ENDIF

    \STATE \textbf{2. Local Parallel Quantization}
    \FOR{row $i$ from $j_{start}$ \textbf{down to} $j_{start}-B$}
        \STATE Update center $c_{:,i}$ using local neighbors within block $J$
        
        \STATE \textbf{Path Diversification (Vectorized Operation):}
        \STATE Path $0$ (Greedy): $\mathbf{q}_{0, i} \leftarrow \text{Round}(\mathbf{c}_{0, i})$
        \STATE Path $1..K$ (Stochastic): $\mathbf{q}_{k, i} \leftarrow \text{Sample}(\mathbf{c}_{k, i}, \text{top-}K)$
    \ENDFOR
\ENDFOR

\RETURN $\mathbf{Q}$

\end{algorithmic}
\end{algorithm}
\begin{algorithm}[h] 
\caption{Quantization with \emph{JTA} with random} \label{a:babai_r}
\begin{algorithmic}[1]
\STATE Compute ${\mathbf{s}}$ and $\mathbf{z}$

\STATE Compute the Cholesky factorization: 
$\widetilde{\mathbf{X}}^\top\widetilde{\mathbf{X}} + \lambda^2\mathbf{I} =\mathbf{R}^\top\mathbf{R}$
\STATE Solve the lower triangular system $\mathbf{R}^\top\mathbf{u}=\mathbf{A}^\top\mathbf{y}$
and the upper triangular system $\mathbf{R}\mathbf{v}=\mathbf{u}$.
\STATE Compute $\bar{\mathbf{q}}=\mathbf{v}\oslash\mathbf{s} +\mathbf{z}$
\STATE Compute $\bar{\mathbf{R}}=\mathbf{RD}$
\STATE $\mathbf{c}(m) = \hat{\mathbf{q}}(m)$  
\STATE $
    \Pr(\mathbf{q}_m=v)\;=\;
    \frac{\exp\!\left(-\alpha\,\bar{\mathbf{R}}_{ii}\,(\mathbf{c}_m- v)^2\right)}
    {\sum_{u=0}^{|\mathbb{B}|} \exp\!\left(-\alpha\,\bar{\mathbf{R}}_{ii}\,(\mathbf{c}_i- u)^2\right)},
    \quad v\in\mathbb{B}$
% \STATE $\mathbf{q}(m)= \text{clamp}(\lfloor \mathbf{c}(m) \rceil; \mathbb{B})$  
\FOR{$i=m-1:-1:1$}  
\STATE $\mathbf{c}(i) = \bar{\mathbf{q}}(i) + \left(\sum_{j=i+1}^m \bar{\mathbf{R}}(i,j)(\bar{\mathbf{q}}(j)-\mathbf{q}(j))\right)/\bar{\mathbf{R}}(i,i)$ \\
\STATE $
    \Pr(\mathbf{q}_i=v)\;=\;
    \frac{\exp\!\left(-\alpha\,\bar{\mathbf{R}}_{ii}\,(\mathbf{c}_i- v)^2\right)}
    {\sum_{u=0}^{|\mathbb{B}|} \exp\!\left(-\alpha\,\bar{\mathbf{R}}_{ii}\,(\mathbf{c}_i- u)^2\right)},
    \quad v\in\mathbb{B}$ \\
\ENDFOR
\end{algorithmic}
\end{algorithm}
\begin{algorithm}[h]
\caption{K-Best Randomized JTA Quantization}
\label{a:kbest_jta}
\begin{algorithmic}[1]
\REQUIRE Input matrices $\widetilde{\mathbf{X}}, \mathbf{y}, \mathbf{W}$, regularization $\lambda$, bit-width wbit, number of trials $K$
\ENSURE Quantized vector $\mathbf{q}^\star$

\STATE Initialize best residual $r^\star \leftarrow +\infty$
\STATE Initialize $\mathbf{q}^\star \leftarrow \mathbf{0}$

\FOR{$k = 1$ to $K$}
    \STATE Run Algorithm~\ref{a:babai_r} to obtain a candidate $\mathbf{q}^{(k)}$
    \STATE Compute residual
    \[
        r^{(k)} = \left\| \mathbf{A}\mathbf{D}\mathbf{q}^{(k)} - \mathbf{b} \right\|_2^2
    \]
    \IF{$r^{(k)} < r^\star$}
        \STATE $r^\star \leftarrow r^{(k)}$
        \STATE $\mathbf{q}^\star \leftarrow \mathbf{q}^{(k)}$
    \ENDIF
\ENDFOR

\STATE \textbf{return} $\mathbf{q}^\star$
\end{algorithmic}
\end{algorithm}
We record the per-layer quantization time on LLaMA3-8B for different values of $K$. 
\begin{figure}
    \centering
    \includegraphics[width=\linewidth]{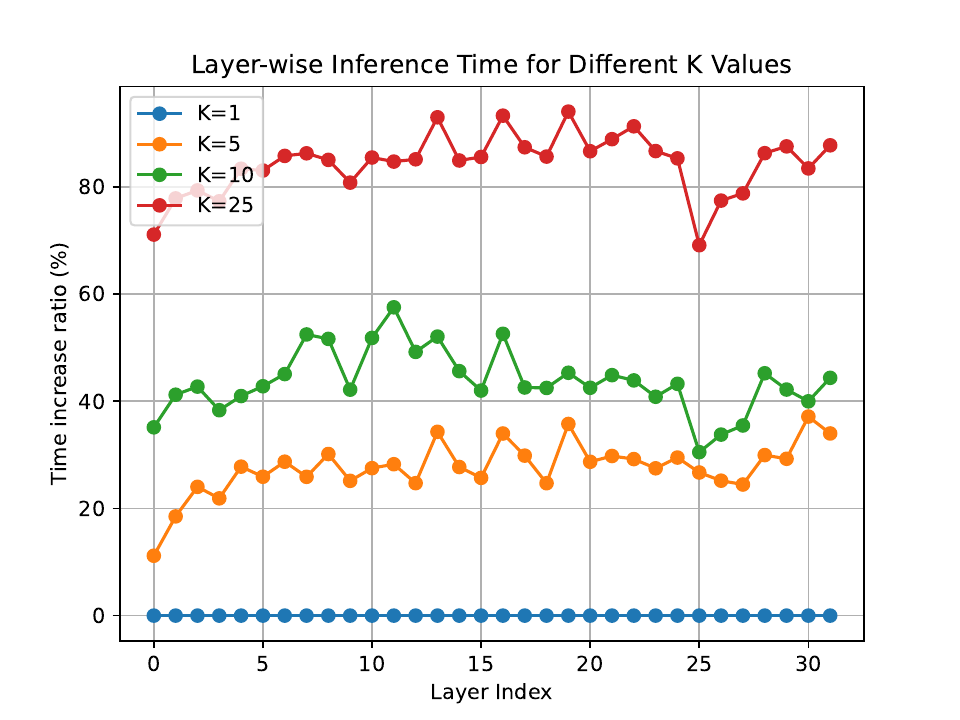}
    \caption{Layer Time increase ratio for different K on Llama3-8B 4 bits}
    \label{fig:time_increase}
\end{figure}
Figure~\ref{fig:time_increase} reports the relative increase in layer-wise computation time under the K-best strategy.

As shown in Fig.~\ref{fig:time_increase}, even when using $K=25$—corresponding to 25 independent randomized decoding paths—the additional computational overhead is modest, amounting to approximately an 80\% increase in per-layer runtime.It suggests the usefulness of our parallelized algorithm

Table~\ref{tab:mu_lambda_grid} presents an ablation over the hyperparameters $\mu$ and $\lambda$ on WikiText-2. We observe that intermediate values of both parameters consistently yield lower perplexity, while overly small or large settings degrade performance. Although the global minimum is attained at $(\mu=0.6,\lambda=0.4)$, the configuration $(\mu=0.6,\lambda=0.6)$ achieves comparable performance (PPL 7.87) and lies within a stable low-perplexity region. For this reason, we fix $\mu=0.6$ and $\lambda=0.6$ in all main experiments, as it provides a robust trade-off without relying on the single best hyperparameter combination.
\begin{table}[t]
\centering
\caption{Perplexity on WikiText-2 for LLaMA-3-8B under different $(\mu,\lambda)$ and 3 bits settings. Lower is better.}
\label{tab:mu_lambda_grid}
\setlength{\tabcolsep}{4pt}
\resizebox{\columnwidth}{!}{
\begin{tabular}{c|cccccccc}
\toprule
$\mu \backslash \lambda$
& 0.1 & 0.2 & 0.3 & 0.4 & 0.5 & 0.6 & 0.7 & 0.8 \\
\midrule
0.1 & 8.2499 & 8.1392 & 8.0821 & 8.0682 & 8.0099 & 8.0259 & 8.0285 & 8.0134 \\
0.2 & 8.1045 & 8.0339 & 7.9979 & 7.9550 & 7.9797 & 7.9594 & 7.9868 & 7.9749 \\
0.3 & 7.9992 & 7.9873 & 7.9506 & 7.9348 & 7.9198 & 7.9590 & 8.0508 & 7.9643 \\
0.4 & 7.9634 & 7.9326 & 7.8940 & 7.9097 & 7.8870 & 7.9471 & 8.0463 & \textbf{7.8804} \\
0.5 & 7.9515 & 7.9277 & 7.8604 & 7.8695 & 8.0539 & 7.8604 & 8.1135 & 7.9339 \\
0.6 & 7.9111 & 7.8839 & 8.0664 & \textbf{7.8404} & 7.9041 & 7.8726 & 8.0893 & 7.8665 \\
0.7 & 7.9198 & 7.8892 & 7.8613 & 7.8656 & 7.8874 & 7.8443 & 8.3100 & 8.2792 \\
0.8 & 7.9172 & 7.8927 & 7.9665 & 7.9049 & 7.8874 & 7.8905 & 8.4475 & 8.3451 \\
0.9 & 8.0570 & 7.9295 & 7.9815 & 7.9594 & 7.9616 & 8.0790 & 8.5767 & -- \\
1.0 & 8.2239 & 8.1054 & 8.1319 & 8.1207 & 8.0602 & 8.0781 & 8.8646 & -- \\
\bottomrule
\end{tabular}
}
\end{table}